# Towards Transparent and Accurate Diabetes Prediction Using Machine Learning and Explainable Artificial Intelligence

Pir Bakhsh Khokhar, Viviana Pentangelo, Fabio Palomba, Carmine Gravino
Software Engineering (SeSa) Lab, Department of Computer Science, University of Salerno, Salerno, Italy
E-mail: {pkhokhar, vpentangelo, fpalomba, gravino}@unisa.it

*Abstract*—Diabetes mellitus (DM) is a global health issue of significance that must be diagnosed as early as possible and managed well. This study presents a framework for diabetes prediction using Machine Learning (ML) models, complemented with eXplainable Artificial Intelligence (XAI) tools, to investigate both the predictive accuracy and interpretability of the predictions from ML models. Data Preprocessing is based on the Synthetic Minority Oversampling Technique (SMOTE) and feature scaling used on the Diabetes Binary Health Indicators dataset to deal with class imbalance and variability of clinical features. The ensemble model provided high accuracy, with a test accuracy of 92.50% and an ROC-AUC of 0.975. BMI, Age, General Health, Income, and Physical Activity were the most influential predictors obtained from the model explanations. The results of this study suggest that ML combined with XAI is a promising means of developing accurate and computationally transparent tools for use in healthcare systems.

## I. Introduction

Diabetes is a global healthcare issue with major prevalence increasing and millions living with the disease [5]. In 2020, type-2 diabetes was among the top ten causes of death, as listed by the World Health Organization [30]. Diabetes mellitus also warrants particular comment due to its chronic nature and many associated complicating conditions, including cardiovascular disease, neuropathy, retinopathy and kidney failure [11], [15], [20]. As a result, prediction of early and accurate onset of diabetes is essential to improve patient outcome and quality of life [25]. Interventions started earlier can help prevent serious complications of the disease in high-risk individuals.

The use of Machine Learning (ML) models to predict and diagnose diabetes from true patient data has been exploited widely [3], [4], [25]. While such models have achieved good predictive performance in predicting diabetes with high accuracy [3], [4], the interpretability of these models has rarely been addressed. Predictions made using complex algorithms are frequently challenging for healthcare professionals to comprehend, limiting their practical applicability in clinical settings. Thus, research has begun to employ eXplainable Artificial Intelligence (XAI) techniques [19], [21] to enhance the transparency of ML predictions. Such increased transparency and interpretability enable healthcare providers, particularly those who are not experts in AI, to interpret, trust, and rely more confidently on the outputs of predictive models.

However, despite significant advancements in XAI techniques and their growing application in interpreting ML model predictions for diabetes, these methods are often treated as isolated black-box tools, primarily focusing on evaluating the results of individual ML algorithms [17], [24], [27]. Indeed, we identified a research gap in the lack of comprehensive empirical studies evaluating both ML and XAI methods in the context of diabetes mellitus prediction. Most existing studies are limited in scope and typically rely on limited datasets [27] or focus on a single XAI algorithm [6], [15], which constrains the generalizability of their findings. Furthermore, no considerations are made for XAI algorithms, nor are any assessments provided on their performance and suitability for explaining health predictions. These limitations highlight the need for a more extensive evaluation of ML and XAI techniques in this domain.

The present study aimed to address the identified research gap by conducting a large-scale empirical analysis on a diverse set of ML models, including both simple and ensemble approaches tailored for diabetes mellitus prediction. We evaluated ML models based on well-established performance metrics and applied multiple XAI techniques to interpret their predictions. Moreover, we performed a detailed comparative analysis not only on the models themselves but also on the XAI methods. By leveraging the comprehensive Diabetes Binary Health Indicators dataset, derived from the Behavioral Risk Factor Surveillance System (BRFSS) and available at Kaggle[1], our ultimate objective is to generate in-depth insights into the effectiveness and usability of both the models and their interpretability techniques, ultimately enhancing the generalizability and accessibility of results for medical practitioners who are not AI experts.

## II. Literature Review

The prediction of diabetes using ML techniques has been a subject of considerable interest, particularly in recent years [3]. However, in the medical context, achieving accurate predictions and identifying the most influential factors to provide valuable explanations and facilitate a deeper understanding of the data and models [26] is crucial. In this regard, techniques aimed at enhancing the clarity and transparency of models, namely XAI techniques, are being investigated to explore their potential for diagnosing and elucidating diabetes.

---
[1]Diabetes Binary Health Indicators:https://www.kaggle.com/datasets/alexteboul/diabetes-health-indicators-dataset

In this section, we present an overview of the state of the art concerning the application of XAI techniques to diabetes prediction, the most frequently considered settings, and the primary insights obtained. Detailed information from related literature is presented in Table I.

Local Interpretable Model-agnostic Explanations (LIME) [21] and SHapley Additive explanations (SHAP) [19] are among some of the ways most often used in the literature. There have been two of these algorithms for producing ML model prediction explanations, and they have become so popular in this context that they serve as the base for many studies in the domain. LIME provides individual prediction explanations rather than the entire model. In particular, it has been used to explain predictions in ML models by detecting key features such as glucose levels [16], [28]. On the other hand, SHAP is already a popular XAI technique based on game theory: it produces both local and global interpretability by distributing feature contributions across all possible combinations. Different models have been utilized to improve clinical decision support systems further and highlight influencing predictors such as HbA1c and glucose levels, with high accuracy, across various models [6]–[8]. SHAP has also been used to find the critical regions in ocular images and to enhance the prediction accuracy [23]; however, ensemble methods such as Eye-Net have shown promising performance in automated screening systems [18].

Most importantly, LIME and SHAP are not mutually exclusive. These have often been combined or compared to realize greater insights into model predictions. Studies integrating both techniques have demonstrated their utility in identifying key features, such as glucose and BMI, enhancing the understanding of global and local model behavior [10], [17], [24]. Comparative analyses also suggested that LIME may offer greater stability in certain cases [1]. The most recent trend in the literature indicates that researchers have begun to explore more comprehensive models and XAI techniques to enhance performance insights [26], which analyzed XAI applications for diabetes prediction, identifying limitations such as poor generalizability from small datasets, and emphasizing the need for improved data quality, diverse real-world sources, and integration with IoT sensors for personalized diabetes management. Vivek et al. [27] compared multiple XAI techniques, including SHAP, LIME, QLattice [2], ELI5 [9], and anchor [22], within a clinical decision support system.

The literature analysis revealed that related work generally treats XAI methods as isolated black-box tools, limiting their application to evaluating ML model outputs alone. Most comprehensive studies either focus on a single XAI technique [6], [7], [16] or apply different techniques without providing insights into their performance [17], [24], [27], highlighting the gap of a systematic evaluation of XAI algorithms for explaining diabetes-related predictions.

> ⚠ **Identified Limitation**
> Most of the literature focuses on individual techniques or models as black boxes without considering interpretability challenges. This leaves questions unanswered, such as identifying effective methods to explain machine learning model outputs, understanding trade-offs between global and local XAI techniques, and evaluating their practical relevance in clinical contexts.

This study is motivated by significant limitations in prior studies on machine learning (ML) and explainable AI (XAI) for predicting diabetes. While fundamental models such as Logistic Regression [6] and SVM [16] have been frequently employed due to their interpretability, they demonstrate inadequate capacity to capture the complex, nonlinear patterns present in structured healthcare information. Ensemble techniques, as investigated by Kibria et al. [17], enhanced accuracy but failed to resolve the interpretability issues crucial for clinical implementation. Additionally, research such as Tasin et al. [24], which combined various XAI approaches, lacked a comprehensive framework to assess the balance between global and local explainability or to incorporate these insights into practical medical workflows.

This investigation aims to fill existing gaps by building upon the state-of-the-art XAI tools, including SHAP, LIME, Explainable Boosting Machines (EBM), and counterfactual explanations, and anchor an ensemble framework. In contrast to prior studies that have targeted these techniques, this study rigorously evaluated these techniques in a healthcare context, highlighting that they can create actionable insights and facilitate improved clinical decision-making. This study seeks to achieve a balance between predictive accuracy and interpretability by avoiding the pitfalls of previous research that have either exclusively applied XAI methods [24], [27] or is not applicable in practical clinical settings. The integration of ensemble techniques and a comprehensive array of XAI tools ensures a dual focus on accuracy and transparency, establishing a new standard for applying AI in healthcare. This approach combines robust predictive ML models with explainability frameworks and allows clinicians to make decisions based on the data from patients with increased confidence.

## III. Materials and Methods

To address the limitations identified in the existing research and guide our investigation and methodology, we formulated a set of research questions. These questions were designed to tackle key areas of uncertainty regarding explainability techniques and their practical implementation in predicting diabetes mellitus.

> ❓ **RQ1**
> Which ML models guarantee the highest predictive accuracy and simultaneously offer comprehensible results that are useful in diagnosing diabetes mellitus?

TABLE I
SUMMARY OF THE RELATED WORK.

| Reference | Objective | Dataset | Prediction Techniques | XAI Technique |
|---|---|---|---|---|
| Khan and Meehan (2021) [16] | Interpret the predictions of diabetes on different ML models | PIMA Indian Diabetic Dataset | SVM, Logistic Regression, K-NN, Decision Tree, Naive Bayes, Random Forest | LIME |
| Du et al. (2022) [8] | Develop an explainable ML-based clinical decision support system for predicting gestational diabetes | PEARS dataset | Logistic Regression, Random Forest, SVM, AdaBoost, XGBoost | SHAP |
| Liu and Cuadros (2022) [18] | Develop an interpretable ML ensemble system for classification and lesion localization of diabetic retinopathy | APTOS and DDR datasets | 24 CNN architectures | Grad-CAM, SHAP |
| Shakeri et al. (2022) [23] | Identify the areas of an ocular image that contribute the most to diabetic retinopathy prediction | APTOS Dataset | ResNet50 | SHAP |
| Kibria et al. (2022) [17] | Improve diabetes prediction using an ensemble method with XAI | PIMA Indian Diabetic Dataset | Neural Network, Random Forest, SVM, Logistic Regression, AdaBoost, XGBoost | SHAP, LIME |
| Wang et al. (2023) [28] | Propose a systematic approach to enhance the explainability of AI for diabetes diagnosis | Unspecified dataset and synthetic data | Neural Network approximated to a simpler classification and regression tree | LIME |
| Das and Ahmed (2023) [6] | Develop an explainable ML approach for diabetes detection using influential features | Dr John Schorling's diabetes dataset [29] | SVM, K-NN, and Naive Bayes | SHAP |
| Tasin et al. (2023) [24] | Develop an automatic diabetes prediction system with a focus on XAI | PIMA Indian Diabetic Dataset | Decision Tree, SVM, Random Forest, Logistic Regression, K-NN, XGBoost, AdaBoost | SHAP, LIME |
| Ganguly and Singh (2023) [10] | Using XAI techniques to study the correlation between the features of diabetic and non-diabetic individuals | PIMA Indian Diabetic Dataset | N/A | SHAP, LIME |
| Dharmarathne et al. (2024) [7] | Develop a self-explainable interface for diagnosing diabetes using ML techniques | PIMA Indian Diabetic Dataset | Decision Tree, K-NN, SVC, XGBoost | SHAP |
| Vivek Khanna et al. (2024) [27] | Develop an XAI-driven clinical decision support system for predicting gestational diabetes | Open-source dataset from Kaggle consisting of 133 pregnant women [14] | Logistic Regression, Decision Tree, Random Forest, SVM, K-NN, AdaBoost, XGBoost, ExtraTrees, LightGBM, CatBoost | SHAP, LIME, ELI5, Qlattice, Anchor |
| Ahmed et al. (2024) [1] | Provide a comparative analysis of LIME and SHAP for explaining ML models in diabetes prediction | Survey responses from the CDC's Behavioral Risk Factor Surveillance System | Logistic Regression, Random Forest | SHAP, LIME |

**? RQ2**
Which types of explanation methods are the most useful in explaining the result of the chosen ML models in diabetes prediction?

**? RQ3**
What are the advantages and drawbacks of combining two or more explainability approaches to improve both global and local interpretability of model results?

**? RQ4**
How to evaluate the quality and reliability of explainability method in ML models and how those metrics affect the interpretability and trustworthiness of predictions?

To address these RQs, we followed a detailed methodology (summarized in Fig. 1) by using the Diabetes Binary Health Indicators dataset and implementing the range of preprocessing steps, model selection, and explainability approaches. The aim was to overcome the existing deficit between model accuracy and the ability to understand the results. This framework offers a clear direction for systematically assessing ML and XAI techniques to enhance their applicability in diabetes prediction.

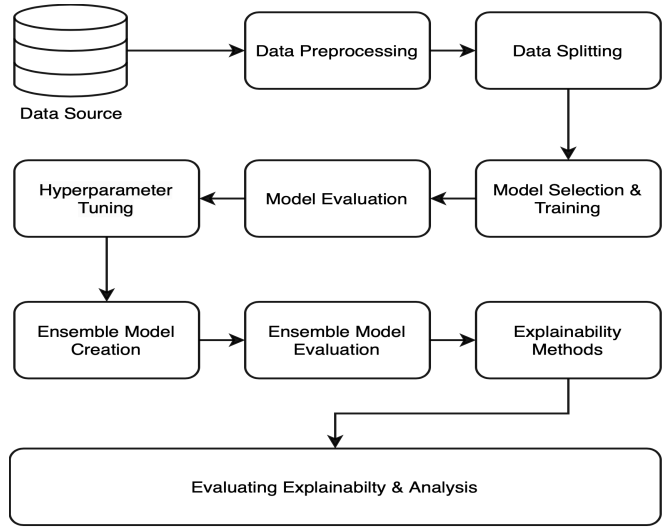

Fig. 1. Proposed Methodology Diagram for Explainable Diabetes Prediction Using Machine Learning

*A. Data Source and Data Preprocessing*

The Diabetes Binary Health Indicators dataset, obtained from the BRFSS, included 253,680 patients with 22 numeric attributes of clinical risk factors, including BMI, cholesterol, blood pressure, activity level, and smoking status. The se-lection variable Diabetes_binary categorizes the population into Diabetic and Non-diabetic, making binary classification possible. However, the database skewed heavily towards class 0, which was non-diabetic, with only 13.07% of the cases being diabetic class one. This imbalance is a challenge in the process of training the model, as it would allegedly be able to predict non-diabetic cases. To tackle this, SMOTE was used on the minority class to synthesize new samples and to train the model independently, i.e., decisional bias towards the majority class might otherwise affect the overall performance.

Data preprocessing makes it possible to feed the dataset into the model through activities such as handling missing values, scaling features, and handling class imbalance. Median imputation handles missing values because this method appropriately assesses healthcare datasets when outliers are frequently noted. When using the median, it retains this central tendency without skewing by large data points or what is referred to as 'outliers'.

Subsequently, feature scaling was done using StandardScaler, which scales each feature to have zero mean and unit variance. This step was needed because health indicators include BMI, blood pressure, and cholesterol, wherein the units will not be the same and there will be problems with scaling. Standardization also enabled every feature to be independent when contributing to the model convergence, reducing the time to convergence and enabling comparability of feature weights.

### B. Data Splitting for Model Evaluation

For model training, tuning, and assessment, the entire dataset was split into training (70%), validation (15%), and test (15 %). Such a split enabled the training set for a pattern, a cross-validation set to tune hyperparameters, and a test set to measure the model's performance, providing a real-world indication of overfitting without compromising generality.

By ensuring that the class distribution was maintained across all subsets of stratified sampling (86.07% non-diabetic and 13.93% diabetic), the real-life split was maintained to support metric evaluations.

### C. Model Selection, Training and Hyperparameter Tuning

Several ML models were built using Random Forest (RF), XGBoost, LightGBM, SVM, Decision Tree (DT), Naive Bayes, and Logistic Regression and tested for the diagnosis of diabetes. These techniques were chosen because they can operate with binary data and yield good results when working with structured health data. All the built models were first trained on the balanced training set created from the original training set, and then the performance was assessed based on the validation set.

Furthermore, we used 3-fold cross-validation and Randomized SearchCV to tune the hyperparameters necessary for each model. This approach allowed a stable tuning technique by avoiding overfitting of the models and guaranteeing that the frequency within all data folds was well-tuned.

### D. Ensemble Model Creation

Ensemble models were built using Random Forest, XGBoost, and LightGBM to ensure robustness in structured health data analysis. The former provides stability and avoids overfitting through bagging [12], [13], whereas XGBoost handles nonlinear relationships efficiently with gradient boosting. LightGBM processes large datasets with a high accuracy in less time.

Soft voting was performed to take advantage of the strengths of these models by averaging their prediction probabilities to allow balanced and robust outputs. Unlike hard voting, which works on the principle of majority decisions, soft voting considers model confidence; hence, it is better for handling class imbalances and weak patterns in diabetes datasets. Optimized precision, stability, and speed of handling such complexities in diabetes prediction.

### E. Model Evaluation

The performance of both individual ML models and the ensemble model was rigorously evaluated using a comprehensive set of metrics: accuracy, ROC-AUC, precision, recall, and F1-score. These metrics offer a thorough evaluation and comparison of ML models.

To avoid overfitting, an evaluation was performed on both the validation and test sets. In addition, the process highlights the need for an ensemble model and for evaluating the performance of ensemble models versus individual models in terms of predictive accuracy and confidence.

### F. Explainability of Models

Numerous explainability approaches have been employed as part of model processing for transparency and to interpret the findings. SHAP offers both global and local feature importance, and thus, it helps us understand which health indicators carry the most significant weight for prediction as well as individual interpretability. Furthermore, a model-specific approach based on feature importance was used to obtain a complete understanding of the impact of the variables on diabetes risk, and critical predictors were identified.

Partial Dependence Plots (PDPs) were used to describe the interaction between features that affected the prediction, i.e., how adding a feature increased the prediction of the outcomes once specific values were chosen for the feature under consideration. For example, PDPs associated with BMI or hypertension have been shown to influence the likelihood of diabetes risk.

For instance-level interpretations, LIME was used in the study because it provides clear heatmaps of the features' contributions to the achieved prediction. The heatmap was a noteworthy feature because it allowed for comparison and explanation of predictions irrespective of the models used in the application. We also considered EBM, which attains high accuracy alongside interpretability by design, and Global Surrogate Models, which replace difficult-to-explicate models with easier-to-explicate ones.

Anchors offered computable rule-based enhancements that stated the conditions under which certain predictions were derived, offered feature-value modifications, and contributed to broadening the understanding of the decision's boundaries.

Overall, the above techniques provide a clear and robust framework for interpreting model outputs and are fundamental for healthcare, as the models used should be accurate and explainable.

### G. Explainability Metrics

Several explanatory metrics were applied to assess the quality and reliability of the model explanations.

- **Fidelity** is the degree to which an explanation aligns with the model's original predictions, accurately representing the model's behaviour.
- **Faithfulness** evaluates if explanations accurately reflect the model's underlying mechanisms, providing real insights into the model's decision-making process.
- **Sparsity** stresses minimizing features with retained relevance, exposing the most important variables in deciding on predictions.
- **Stability** promotes similar inputs to result in similar explanations, generating more reliable model outputs.
- **Consistency** assesses the degree of agreement in feature importance for different models or conditions as its main value for validating results in varied situations.

Such measures offer a strong foundation for assessing the quality of model explanations and aligning their interpretability and accuracy in the context of healthcare.

In our study, we developed an accurate diabetes prediction model using integral data preprocessing, multiple modelling and ensembles, and explainability techniques. The model is interpretable and can generate informative predictions for clinicians. The proposed ensemble scheme demonstrates the potential of solutions for healthcare decision-making.

## IV. ANALYSIS OF THE RESULTS

In this section, we present and analyze the results of our study according to the four research questions related to model performance, explainability methods, and the advantages and disadvantages of combining explainability methods and their assessment of explainability metrics.

### A. RQ1 - ML Models for Diabetes Prediction

The highest predictive accuracies were achieved by individual models, such as XGBoost, LightGBM, and RF, because these can model nonlinear relationships and complex interactions in healthcare datasets. These models are robust to variations in data structures and are, therefore, suitable for predicting diabetes mellitus. Surprisingly, the best result among the three models was the ensemble model using soft voting, which achieved a test accuracy of 92.5% and a ROC-AUC of 0.975 (see Table II). Using the ensemble strategy, we used XGBoost for its precision, LightGBM for its speed with large datasets, and Random Forest to reduce overfitting, thus retaining interpretability simultaneously as a generalization. This balance between accuracy and practical utility makes such a combination highly relevant for clinical decision-making.

> ✓ **Answer to RQ1**
> Ensemble model outperformed other ML models in terms of accuracy and ROC-AUC. This combination successfully retains the non-linearity while still being interpretable and thus proved to be the most suitable for diagnosing diabetes mellitus.

### B. RQ2 - Explanation Methods for ML Predictions

We used various explainability techniques to reveal global patterns and explain feature interactions, generate personalized insights to help healthcare professionals understand and trust machine learning predictions, and build a robust framework.

*1) Global Interpretability - Key Predictors Driving Diabetes Risk:* We used global interpretability tools to identify the most important predictors and their overall contribution to diabetes risk. SHAP was used to explain the importance of global and local features. Important drivers contributing to diabetes risk included BMI, GenHlth, Age, and income (see Fig. 2). The results also indicated that BMI level has a positive correlation, and general health and physical activity have a moderating impact, as they act as buffer variables. SHAP enhances the depiction of such associations to facilitate the identification of clinical intervention priorities.

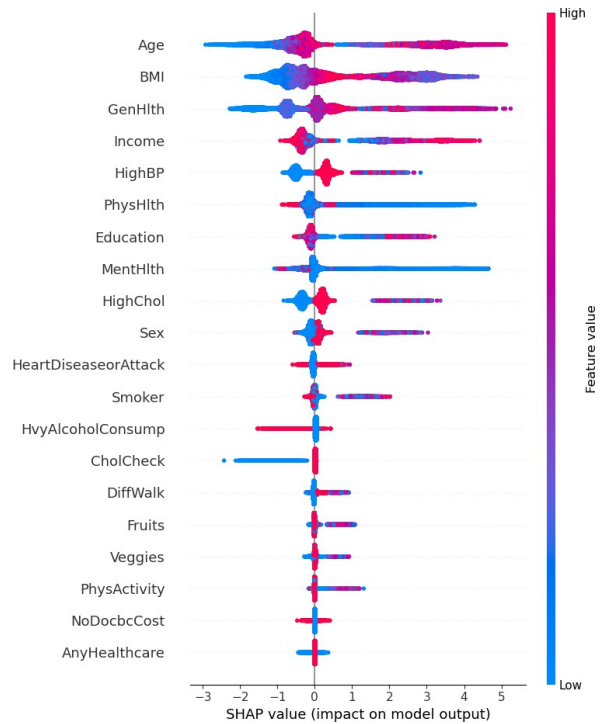

Fig. 2. Key features influence diabetes prediction - SHAP Summary Plot

The results from the EBM indicated that it reached accurate predictions while being comprehensible by clinicians and patients. As shown in Fig. 3, EBM demonstrated the particularity of ranks, where BMI and GenHlth were placed as the highest priority of the most important predictors. EBM's capacity of the EBM to combine accuracy and interpretability by construction increases its effectiveness in solving specific healthcare problems.

TABLE II
COMPARISON OF INDIVIDUAL AND ENSEMBLE MODEL PERFORMANCE.

| Model | Validation Accuracy (%) | Test Accuracy (%) | Validation ROC-AUC | Test ROC-AUC | Precision | Recall | F1-Score |
|---|---|---|---|---|---|---|---|
| XGBoost | 91.62 | 91.81 | 0.97 | 0.97 | 0.97 | 0.86 | 0.91 |
| LightGBM | 91.86 | 92.03 | 0.97 | 0.97 | 0.97 | 0.86 | 0.91 |
| Random Forest | 90.96 | 91.02 | 0.966 | 0.96 | 0.94 | 0.87 | 0.91 |
| Decision Tree | 88.66 | 88.50 | 0.93 | 0.92 | 0.91 | 0.86 | 0.88 |
| SVM | 89.50 | 89.40 | 0.95 | 0.94 | 0.93 | 0.86 | 0.89 |
| Logistic Regression | 75.09 | 74.90 | 0.83 | 0.82 | 0.74 | 0.77 | 0.75 |
| Naive Bayes | 72.26 | 71.80 | 0.79 | 0.78 | 0.72 | 0.72 | 0.72 |
| Ensemble Model | 92.50 | 92.80 | 0.97 | 0.97 | 0.97 | 0.87 | 0.92 |

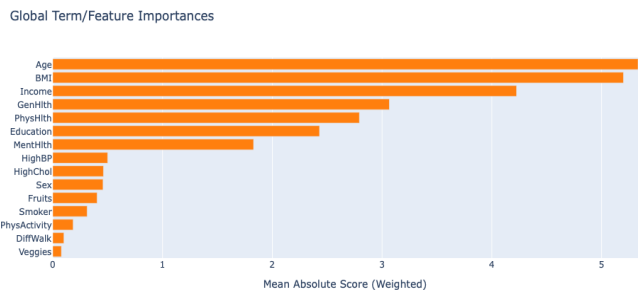

Fig. 3. EBM Feature Importance – Top Global Predictors

SHAP and EBM corroborated the findings of Permutation Importance, which is that BMI, general health, and physical health (PhysHtlth) were the most important predictors of the risk of diabetes. Fig. 4 illustrates how consistently these features ranked highest, achieving stability and consistency with clinical expertise. This consistency shows us that the role played by lifestyle and overall well-being in managing diabetes are factors, that BMI and overall health represent metabolic and endocrine health, and that physical health is important for active intervention. Permutation Importance provides interpretable rankings that increase confidence in the model and decision-making process by providing early detection and personalized healthcare interventions.

*2) Local Interpretability - Personalized Insights for Individual Patients:* Local interpretation helped resolve ambiguity at the micro level and gave a precise idea about some local predictions to help clinicians devise fine-tuned patient treatment plans. LIME provides post hoc, understandable, and concrete examples related to each prediction. As shown in Fig. 5, three features that impacted the prediction of the particular patient were Age, BMI, gender, Income, and Physical Health. This representation disaggregates abstract forecasts and concrete behaviour-change strategies, such as increasing physical activity or lowering blood pressure.

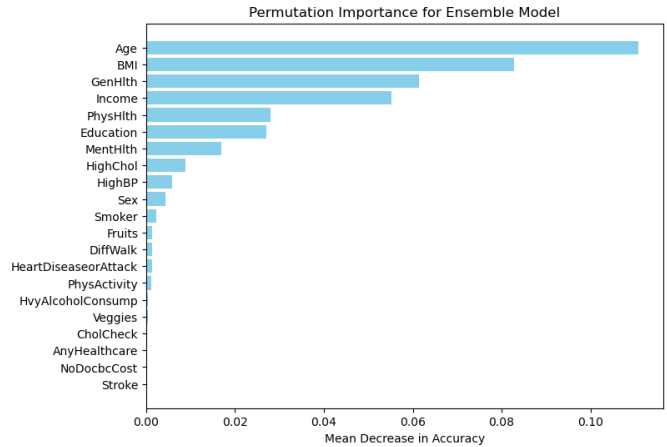

Fig. 4. Ranking of Feature Importance for the Ensemble Model - Permutations Importance

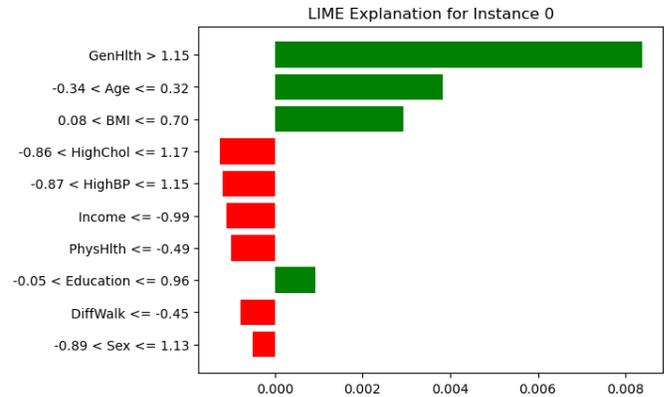

Fig. 5. LIME Explanation for Instance Level Predictions (Instance 0)

The SHAP force plots (Fig. 6) and the waterfall plots (Fig. 7) showed decomposed individual predictions into an additive feature contribution. For instance, BMI and high cholesterol levels were found to have a strong positive relationship with diabetes in a specific patient. Still, features such as income were found to decrease diabetes. These visualizations enable clinicians to easily distinguish the nature of various factors at the patient level.

In addition to the LIME and SHAP force plots, a decision plot (Fig. 8) breaks down the contribution toward a prediction into sequential feature contributions. For instance, Age, BMI, and GenHlth contribute significantly to increasing diabetes risk, whereas income lowers it. These insights can inform clinical decision-making, enhance patient-specific treatment strategies, and improve the overall efficacy of healthcare delivery.

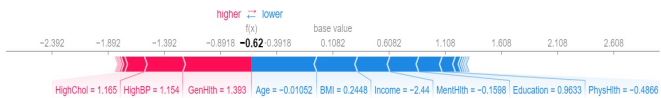

Fig. 6. Positive and Negative Feature Effects for a Single Prediction using SHAP Force Plot

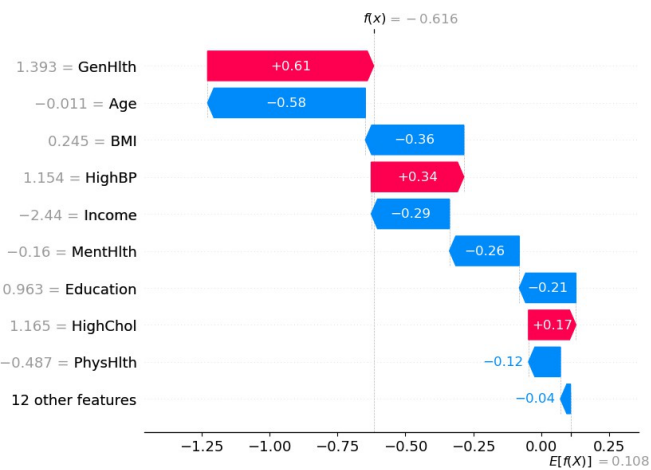

Fig. 7. Feature Contributions Explaining Individual Predictions - Waterfall Plot

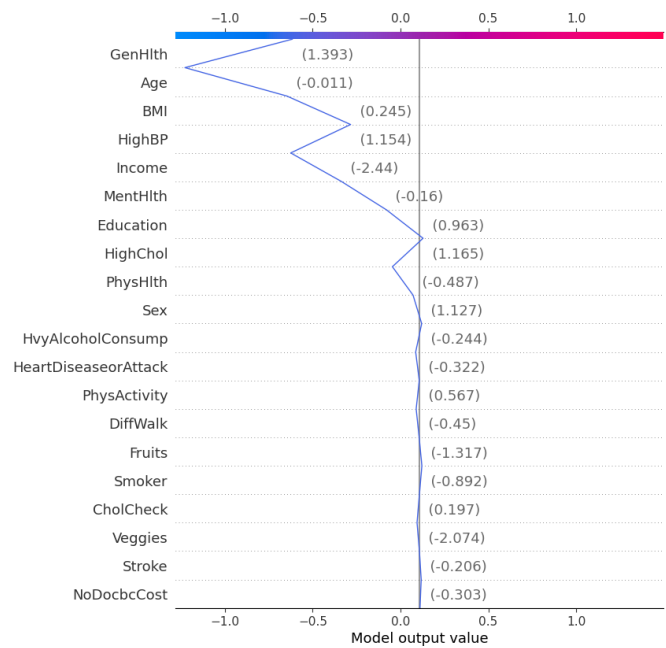

Fig. 8. Cumulative Feature Contributions Impacting Predictions - Decision Plot

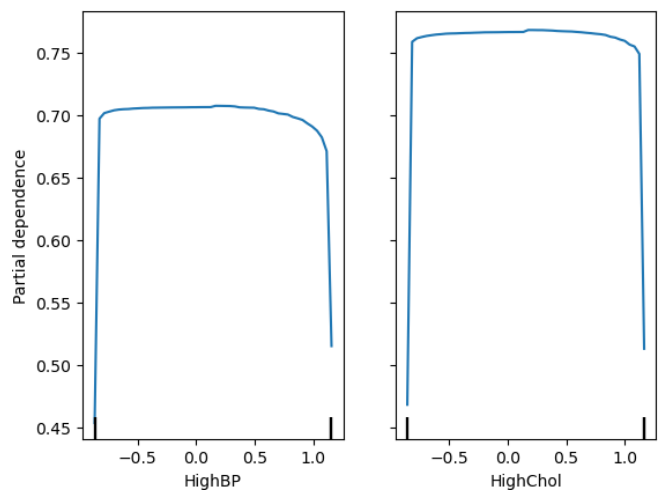

Fig. 9. Partial Dependence Plot – Interaction between blood pressure and cholesterol levels.

*3) Feature Interactions and Contextual Interpretability:* The analysis of the features interaction provided a breakdown of contextual descriptions to expound on how the predictors affected diabetes risk. Understanding the interactions, PDPs explained that there are no simple linear linkages between features; for example, in the PDPs for HighBP and HighChol (Fig. 9), the incidence of diabetes in those with both high blood pressure and high cholesterol levels was extremely high. This is mainly because PDPs are particularly useful in explaining how various feature values combine to affect outcomes.

Explanations for diabetes risk predictions by the anchors were defined and in line with the rules. As shown in Fig. 10, patients with low physical activity and BMI greater than 30 were defined as high risk. These thresholds are easy to interpret and can provide practitioners with immediate guidance on making decisions.

*4) Actionable Recommendations through Counterfactual Explanations:* Actionable detail, shown in Table III, includes counterfactual explanations of "what if" framings of how specific features lower the risk of developing diabetes. These were a decrease in BMI (0.24), an increase in physical activity (0.57), and general health (1.39), which accounted for much of the prediction. These explanations also further demonstrate that controlling factors such as HighBP (coefficient of 1.15) and HighChol (coefficient of 1.17), in addition to bettering dieting and curtailing excessive alcohol intake (-0.24), alter risk classifications. These implications bridge model predictions with clinical experience to provide specific recommendations for diabetes prevention and management. Counterfactual explanations bridge AI predictions to actions that are both interpretable and actionable while providing personalized, proactive diabetes care.

TABLE III
COUNTERFACTUAL EXPLANATIONS FOR DIABETES PREDICTION

| HighBP | HighChol | CholCheck | BMI | Smoker | Stroke | Heart Disease or Attack | Phys Activity | Fruits | Veggies | Hvy Alcohol Consump | Any Health-care | No Doc bc Cost | GenHlth | MentHlth |
|---|---|---|---|---|---|---|---|---|---|---|---|---|---|---|
| 1.15 | 1.17 | -2.89 | 0.24 | -0.89 | -0.21 | -0.32 | 0.57 | -1.32 | -0.32 | -0.24 | 0.23 | -0.30 | 1.39 | -0.16 |
| 1.15 | 1.17 | 0.20 | 0.24 | -0.89 | -0.21 | -0.32 | 0.57 | -1.32 | -2.07 | -0.24 | 0.23 | -0.30 | 1.39 | -0.16 |
| 1.15 | -0.21 | 0.20 | 0.24 | -0.89 | 1.26 | -0.32 | 0.57 | -1.32 | -2.07 | -0.24 | 0.23 | -0.30 | 1.39 | -0.16 |

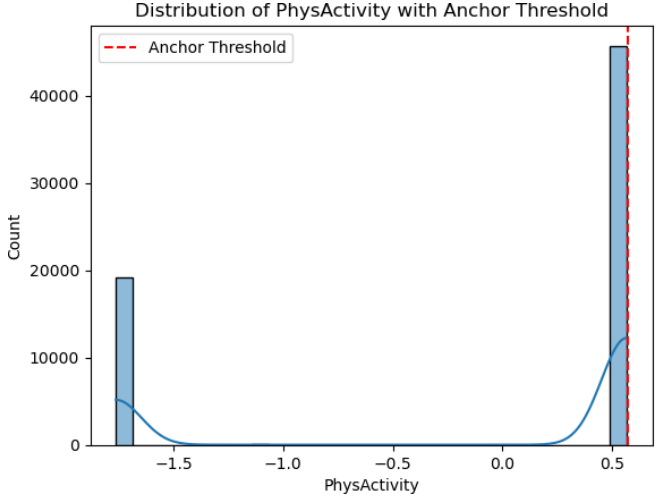

Fig. 10. Anchor Threshold – Physical activity and BMI thresholds for high diabetes risk.

✓ **Answer to RQ2**
SHAP and EBMs fared well in assigning high importance to predictors such as BMI and general health and LIME gave individual focus. Feature interactions were explained between PDPs and Anchors, while Counterfactual Explanations provided potential interventions to build a strong internal structure of diabetes prediction.

*5) RQ3 - Combining Explainability Approaches:* The achieved results prove the value of integrating explainability approaches in interpreting ML models for diabetes prediction. The study strikes a balance of understanding feature importance while still offering personalized insights by combining global methods, such as SHAP and EBM, together with local methods, such as LIME. By highlighting the role of BMI and general health, SHAP elucidates how these factors affect individual predictions. In contrast, LIME helps them understand the specific drivers within each person's predictions to lead them to the right treatment path. Cross-validation and consistent SHAP and EBMs identified BMI, physical activity, and general health as critical predictors. By anchoring, we simplify decision-making while providing intuitive thresholds based on clinical practice. Integration adds complexity; different outputs from methods can overwhelm users or contain contradictory insights. Despite these drawbacks, the combination has the potential to offer a robust trade-off between transparency and actionability for diabetes prediction and management.

✓ **Answer to RQ3**
SHAP, LIME, PDPs, and Anchors enhance global and local interpretability of models by defining important predictors, explaining the prediction of an individual, and describing interactions and decision rules, this comes at the price of complexity and computation, and careful handling is needed for that.

*6) RQ4 - Quality, and Reliability of Explainability Methods:* The evaluation metrics used to assess the quality of explainability methods in diabetes prediction models are provided in Table IV. The fidelity value ensures that the explanations accurately represent the model predictions, and faithfulness of 0.709 confirms that they also depict the model's decision-making. Trust is built by uniformly explaining the model's behavior together. Sparsity makes interpretations easier to understand by following only the most important features, enabling more straightforward application results, such as diabetes prediction. Stability produces consistent explanations for similar inputs, a cornerstone characteristic for clinical applications that is critical for consistency. This validation indicates that key predictors go away inconsistently across models or scenarios, meaning making choices based only on consistency is more generalizable. Together, these metrics pro-vide useful explainable methods for boosting the transparency and trustworthiness of machine learning in predicting diabetes, to be applied in clinical decision support.

TABLE IV
EXPLAINABILITY METRICS FOR THE ML MODEL.

| Metric | Value |
|---|---|
| Fidelity | 0.744 |
| Faithfulness | 0.709 |
| Sparsity | 10.0 |
| Stability | 5.35e-05 |
| Consistency | 0.73 |

✓ **Answer to RQ4**
Explanation accuracy, transparency, and reliability are guaranteed by explanation metrics so that predictions are more interpretable and trustworthy for diabetes diagnosis.

## V. DISCUSSION

The results of this study demonstrate the accuracy, interpretability, importance of explainability metrics, and clinical usability of ML models for diabetes prediction. This section describes more of these insights and provides actionable outcomes for both researchers and practitioners.

*Comparative Analysis with Existing Studies:* In studies using logistic regression, decision trees, and random forests, accuracies were in the range of 70% to 91% and were not robust across a variety of datasets [4], [6], [17]. Leveraging ensemble models and dealing with the imbalances in the data, this study was able to achieve an accuracy of 92.50% and ROC-AUC of 0.975, outperforming previous studies. Results show that ensemble techniques resulted in a dramatic improvement in precision and recall over previous work. Previous studies predominantly used isolated explainability methods, which restrain them from aggregated insights [6], [23]. They lack concrete recommendations or patient-specific insights, such as BMI, age, and glucose levels [10], [24] without considering more general lifestyle factors. This combination of SHAP and LIME revealed key predictors, such as Age, BMI, General Health, and physical activity, leading to actionable recommendations for increasing activity or reducing BMI.

These findings impact both research and practice. This study provides a framework for combining ensemble learning with explainability tools and establishes a new standard for AI healthcare research. It allows actionable insights, such as counterfactual explanations, to help clinicians make evidence-based decisions and propose lifestyle changes. Anchors set thresholds for high-risk classification [23], [27]. This study has added advantages in terms of accuracy, insights, and generalizability. It addresses the gaps in past research and offers a practical framework for enhancing diabetes prediction and care across various healthcare settings.

*Potency of ML Models for Diabetes Prediction:* The ensemble model outperforms previous research due to strategic methodological choices like XGBoost, LightGBM, Random Forest, SMOTE for class balancing, and soft voting for effective aggregation, enabling the model to identify complex feature interactions.

The framework effectively addresses false negatives in diabetes diagnosis by incorporating clinically relevant predictors like BMI, age, and physical activity. This aligns with clinical priorities, enabling actionable recommendations and targeted interventions, bridging the gap between abstract predictions and practical applications. This methodology could be extended to include multimodal data or to validate across other demographics as future research. Model performance is similar to that of human practitioners and can serve as a clinical decision-support tool. It provides transparent insights and greater diagnostic precision, helps reduce clinician burden and improves patient outcomes. AI integration through clinical workflows represents a defining moment for the use of AI in personalized care and equitable medical decision-making.

*Effectiveness of Explainability Methods:* According to the study, integrating several explainability methods (SHAP and EBMs) also increases model transparency and clinical usage. It strengthens the trust between healthcare professionals by showing that the alignment of global predictors like Age, BMI, and general health is in line with clinical priorities. Local methods such as LIME provide patient-specific risk assessments, while Anchors provide rule-based levels of high-risk classification. By demonstrating how BMI and physical activity interact with diabetes risk, PDPs allow clinicians to understand the features of diabetes risk better. Model outputs are translated into actionable steps (e.g. increase physical activity, actions to reduce BMI). In contrast to previous work that leveraged one single explainability approach or a few features [6], [23], this work combined global and local methods. This combination provides translational insights by drawing predictive models and their clinical implementation, and the results are interpretable and impactful in real-world healthcare settings. A combination of a comprehensive and specific approach is used to provide the totality of patient information, including lifestyle compounds like physical activity.

TABLE V
COMPARATIVE ANALYSIS WITH EXISTING STUDIES.

| Aspect | Existing Studies | Current Study |
| --- | --- | --- |
| Predictive Accuracy | The achieved accuracies ranged from 70% to 91%, depending on the dataset and model ( [4], [6], [17]). | A 92.50 % accuracy and ROC-AUC of 0.975 were achieved using advanced ensemble models such as XGBoost and LightGBM. |
| Explainability | It is limited to individual tools such as SHAP for feature importance or ignored entirely ( [6], [23]). | Comprehensive integration of global importance (SHAP), local insights (LIME), rule-based transparency (anchors), and counterfactual explanations (actionable insights). |
| Feature Importance | They primarily focused on BMI, age, and glucose levels ( [10], [24]). | Broadened scope by including lifestyle factors such as physical activity and general health alongside BMI and age. |
| Complexity vs. Transparency | Often, interpretability for accuracy (e.g., ensemble methods), or vice versa (e.g., decision trees) [10]. | Balanced performance and transparency using ensemble models with post-hoc explainability tools (e.g., SHAP, LIME) and inherently interpretable models such as Explainable Boosting Machines (EBMs). |
| Actionability | Limited to feature attribution without actionable recommendations for interventions ( [23], [27]). | It provides actionable insights via counterfactual explanations, enabling clinicians to propose specific changes, such as reducing BMI or increasing physical activity. |
| Generalizability | Relying heavily on specific datasets, such as the Pima Indians Diabetes Dataset, limits its applicability to broader populations [4]. | The proposed external validation of diverse datasets (e.g., NHANES) and advocated multimodal data integration (e.g., combining tabular data with imaging or continuous monitoring). |

*Evaluation of Explainability Metrics:* The explainability metrics showed that machine learning systems can be successfully used in clinical environments. Fidelity and faithfulness measures of outputs from the model are consistent with the internal mechanisms of the model, providing assurance and aiding in making informed medical decisions. Due to the sparsity metric, the interpretation of critical predictors, including BMI and age, becomes more achievable for healthcare professionals. Stability means that the same inputs predict the same explanations, and consistency means that we can build reliable insights from different patient cohorts. Thus, these evaluation criteria indicate that the system has the potential as a reliable clinical decision-support tool. Thanks to the predictions linked to reasonable insights, the model suggests both high-risk patients early on and actionable interventions such as lifestyle changes. By emphasizing explainability and robustness, we set a standard for machine learning in healthcare integration to improve diagnostic precision and patient outcomes while directing further research to develop XAI as a clinical application.

*Advantages and Drawbacks of Combining Explainability Approaches:* Global and local explainability techniques used in conjunction with machine learning enhance predictive capabilities by supplying comprehensive system insights and personalized recommendations. Global methods, such as SHAP, identify critical predictors, such as BMI and exercise habits, while local methods, such as LIME and Anchors, provide patient-specific interpretations. This dual-level strategy enhances clarity, credibility, and clinical relevance but may increase computational power. This study presents a reproducible methodology for incorporating explainability tools into medical settings, offering guidance on lifestyle modifications to reduce the risk of diabetes. It aligns with clinical objectives, enhancing diagnostic precision, health outcomes, and operational efficiency. The model enables healthcare professionals to make informed decisions with confidence and enhances overall medical outcomes.

*Clinical and Broader Implications:* This study assists healthcare providers in understanding predictions through the utilization of SHAP, EBM, and LIME, translating these into personalized recommendations such as lifestyle modifications or monitoring protocols. This methodology aligns with clinical reasoning, enhances confidence, and integrates with electronic health records (EHRs) or real-time monitoring systems, thereby optimizing clinicians' workflows. Our study demonstrates to researchers how the combination of ensemble models with explainability techniques can balance predictive accuracy and transparency, providing a replicable methodology for various medical fields. For practitioners, this framework reduces false negatives, minimizes delayed treatments, and offers accurate, interpretable results tailored to individual patient needs. It supports collaborative decision-making by providing patients with clear risk explanations and actionable measures, thereby fostering trust to medical advice.

On a broader scale, the system has significant public health implications, identifying high-risk populations and informing preventive strategies. Economics, in this case, is about minimizing resource allocation and the financial burden of late-stage disease. The study solves the 'black box' issue for accountability and transparency in building trust between healthcare professionals and patients. It lays down the groundwork for how we can bring ethical AI implementation to light, enhancing diagnostic accuracy and improving patient outcomes while making life easier for healthcare providers. The system takes AI-driven personalized medicine to a new level, incentivizing further work and its immediate adaption into healthcare to give the best outcomes with transparency and assurance.

*Limitations and Future Directions:* Further research and practical application would require a few limitations of this study to be addressed. In reality, artificial data for balancing classes might not reveal real-world challenges, and other studies should widen its scope of data augmentation methods. However, the ligand model cannot be applied technically in the clinical setting due to its large computational requirements. Reducing computational costs and expanding the model to include multiple data types (e.g. imaging or continuous monitoring) will be needed to enhance its clinical value further. Moreover, we will include social factors such as income, education, and the availability of healthcare services that increase model relevance in other demographic groups. Other diabetes research should focus on adapting the model for various patient populations and healthcare settings alike, making it useful in managing diabetes in different settings more broadly.

VI. CONCLUSION

This study demonstrates how various machine learning algorithms and explainable methods can lead to diabetes prediction based on BMI, general health conditions, and physical activity level. These models exhibit high accuracy and reproducibility using both global and local explainability methods, thus providing clear guidance for clinical decisions. The practical insights generated by these models may enhance targeted treatments and optimize clinical practices, rendering these models valuable for clinicians. From a scientific perspective, this investigation necessitates further validation using clinical information from real cases, as well as advancements in improving the interpretability of the models. Interoperability with electronic health records and continuous monitoring systems is essential. Consequently, it will facilitate the development of more accurate, sensitive, and individualized methods and approaches to care and expand the application of artificial intelligence in the field of medicine. In practical terms, the results emphasized how the establishment of trust through explainable artificial intelligence is critical when implementing models in clinical practice, as other healthcare professionals can comprehend the outputs that the model provides, as well as gain confidence in applying them for patient care. The results establish a foundation for the advancement of a preventive healthcare system, appropriate diagnostic tools, and improved patient care. In the future, there is a need to enhance the flexibility and generalizability of these models to play key roles in redesigning telemedicine and the future of efficient health services provision and treatment.

ACKNOWLEDGMENT

The authors thank the Center for Disease Control and Prevention's Behavioral Risk Factor Surveillance System (BRFSS) for providing access to the Diabetes Binary Health Indicators dataset, which was integral to this study. This work has been partially supported by the European Union through the Italian Ministry of University and Research, Project PNRR "D3-4Health: Digital Driven Diagnostics, prognostics and therapeutics for sustainable Health care". PNC 0000001. CUP B53C22006090001